\documentclass[11pt]{llncs}
\usepackage[margin=1in]{geometry}
\usepackage{times}  
\usepackage{helvet}  
\usepackage{courier}  
\usepackage[hyphens]{url}  
\usepackage{graphicx} 
\usepackage{natbib}
\usepackage{tikz}
\usetikzlibrary{positioning,arrows.meta,calc}

\usepackage{caption}
\usepackage{algorithm}
\usepackage{algorithmic}
\usepackage{newfloat}
\usepackage{listings}
\DeclareCaptionStyle{ruled}{labelfont=normalfont,labelsep=colon,strut=off} 
\floatstyle{ruled}
\newfloat{listing}{tb}{lst}{}
\floatname{listing}{Listing}
\title{Learning-Augmented QUBO for Time-Expanded Multi-Commodity Flow}

\author{Aikaterini Nikolidaki}
\authorrunning{Aikaterini Nikolidaki}
\institute{School of Electrical and Computer Engineering\\ National Technical University of Athens.\\
\email{aiknikol@mail.ntua.gr}}

\usepackage{graphicx}
\usepackage{amsmath}
\usepackage{mathtools}
\usepackage{amssymb}
\usepackage{subfig}
\usepackage{booktabs}
\usepackage{float}

\usepackage{xspace}

\usepackage{xcolor}

\def\keywords#1{\par\noindent{\small\textbf{Keywords:}#1}}

\begin{document}

\maketitle

\begin{abstract}
We study energy-aware multi-commodity flow on time-expanded graphs, a problem that captures dynamic routing in layered networks and is NP-hard in general. We formulate the problem as a quadratic unconstrained binary optimization problem, give an explicit penalty coefficient that provably guarantees feasibility, and show that it is smaller than the standard bound by a factor of roughly the number of active links. The layered structure of the time-expanded graph yields three structural results: a treewidth bound that gives a fixed-parameter tractable exact algorithm in the bag size, a matrix-product-operator representation of small bond dimension, and a proof that the annealing gap of the associated Ising Hamiltonian is strictly positive for every instance with a unique optimum, uniformly over bounded windows. For learning-augmented routing, we bound the Hamming distance between a predicted and an optimal flow by the flow error, and prove a consistency guarantee: a hybrid algorithm that keeps the better of the prediction and the solver output returns a flow whose cost exceeds the optimum by a factor that grows linearly with the prediction error. The results are exact statements about the QUBO and its associated Ising Hamiltonian, and they hold for any physical implementation.
\end{abstract}

\keywords{Time-expanded graphs; multi-commodity flow; quadratic unconstrained binary optimization; treewidth; quantum annealing; learning-augmented algorithms; LEO satellite networks}

\section{Introduction}
\label{sec:intro}
Multi-commodity flow is one of the most studied problems in combinatorial optimization, and its integral variant is NP-hard even for two commodities~\cite{even1976}. In many applications, however, the network is not static: links appear and disappear over time, and a routing decision must specify not only \emph{which} path a commodity takes but \emph{when} it takes it. The standard abstraction for such problems is the \emph{time-expanded graph}, in which each time step is a layer, each node is replicated across layers, and edges connect consecutive layers. Time-expanded multi-commodity flow captures a wide range of applications, from dynamic traffic assignment to delay-tolerant networking.
In this paper we study a variant that we call \emph{energy-aware multi-commodity flow} (EA-MCF), in which every edge carries both a capacity and an energy cost, and the objective is to route all commodities at minimum total energy. The formulation is purely combinatorial: the time-expanded graph is an abstract layered DAG, and the energy costs are non-negative integers. Our first observation is that EA-MCF on the time-expanded graph is NP-hard even for two commodities and a horizon proportional to the number of edges, by reduction from integral two-commodity flow on acyclic digraphs (Theorem~\ref{thm:nphardness}). This sets the complexity-theoretic baseline for everything that follows.

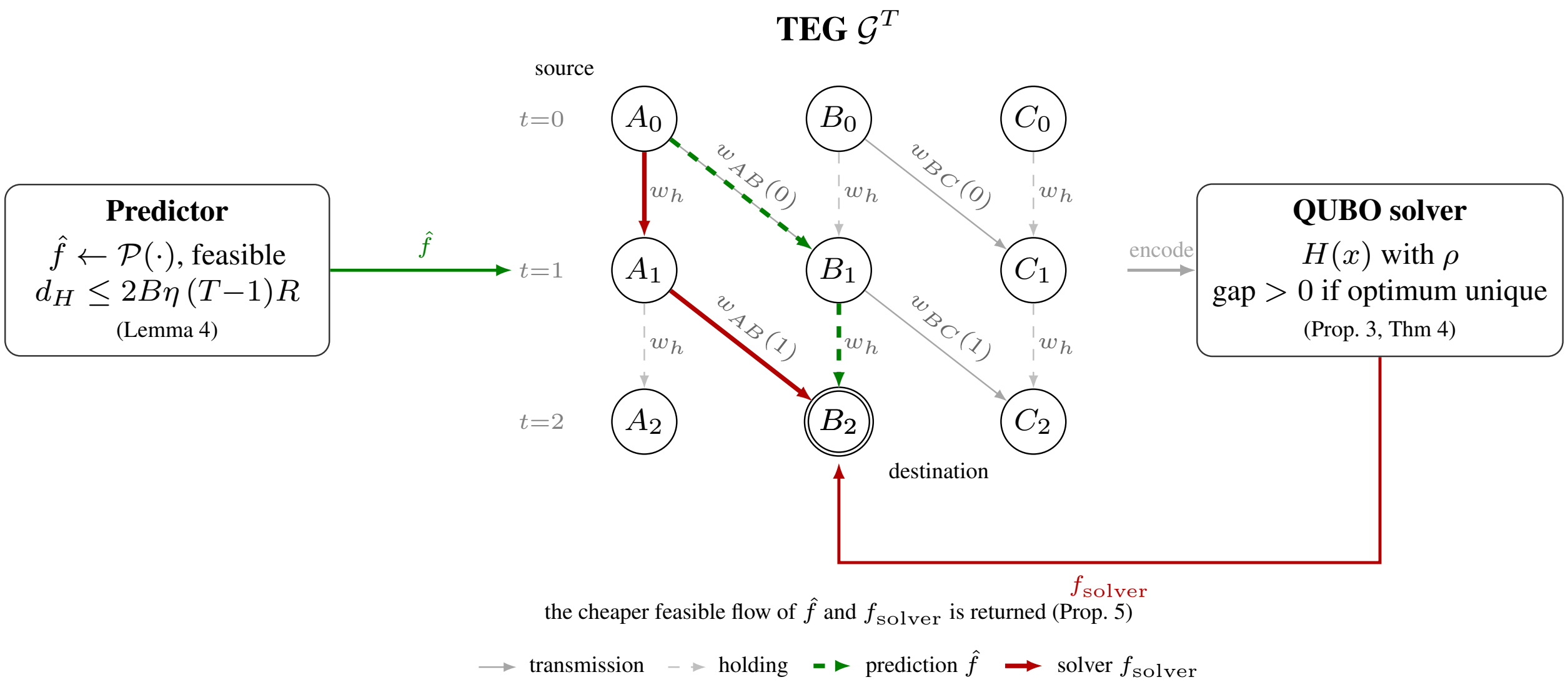
\begin{figure}[t]
\centering
\resizebox{\linewidth}{!}{%
\begin{tikzpicture}[
 sat/.style={circle, draw, minimum size=6mm, inner sep=0pt, font=\scriptsize},
 trans/.style={-{Latex[length=1.2mm]}, thin, gray!70},
 hold/.style={-{Latex[length=1.2mm]}, thin, dashed, gray!50},
 predpath/.style={-{Latex[length=1.5mm]}, very thick, dashed, green!50!black},
 solvepath/.style={-{Latex[length=1.5mm]}, very thick, red!70!black},
 toolbox/.style={draw=black!80, rounded corners, align=center, inner sep=4pt,
                 font=\scriptsize, minimum width=3.0cm},
 flowarrow/.style={-{Latex[length=1.5mm]}, thick, gray!70},
 cost/.style={font=\tiny, gray!80!black}
]
\begin{scope}
 \node[font=\tiny, gray] at (-0.95, 0)    {$t{=}0$};
 \node[font=\tiny, gray] at (-0.95,-1.4)  {$t{=}1$};
 \node[font=\tiny, gray] at (-0.95,-2.8)  {$t{=}2$};
 \foreach \t/\y in {0/0, 1/-1.4, 2/-2.8} {
   \node[sat] (A\t) at (0,\y)   {$A_{\t}$};
   \node[sat] (B\t) at (1.8,\y) {$B_{\t}$};
   \node[sat] (C\t) at (3.6,\y) {$C_{\t}$};
 }
 \node[sat, double, double distance=0.6pt] at (1.8,-2.8) {$B_2$};
 \foreach \t/\tp in {0/1, 1/2} {
   \draw[trans] (A\t) -- node[sloped, above, pos=0.5, cost] {$w_{AB}({\t})$} (B\tp);
   \draw[trans] (B\t) -- node[sloped, above, pos=0.5, cost] {$w_{BC}({\t})$} (C\tp);
 }
 \foreach \nm in {A, B, C} {
   \draw[hold] (\nm0) -- node[right, xshift=-2pt, cost] {$w_h$} (\nm1);
   \draw[hold] (\nm1) -- node[right, xshift=-2pt, cost] {$w_h$} (\nm2);
 }
 \draw[predpath] (A0) -- (B1);
 \draw[predpath] (B1) -- (B2);
 \draw[solvepath] (A0) -- (A1);
 \draw[solvepath] (A1) -- (B2);
 \node[font=\small\bfseries] at (1.8, 0.85) {TEG $\mathcal{G}^T$};
 \node[font=\tiny, anchor=east] at (-0.35,0.45) {source};
 \node[font=\tiny, anchor=west] at (2.15,-3.25) {destination};
\end{scope}
\node[toolbox, left=26mm of A1] (learn) {
 \textbf{Predictor}\\[2pt]
 $\hat{f}\gets\mathcal{P}(\cdot)$, feasible\\
 $d_H\le 2B\eta\,(T{-}1)R$\\
 {\tiny(Lemma~\ref{lem:hamming})}
};
\node[toolbox, right=30mm of B1] (quant) {
 \textbf{QUBO solver}\\[2pt]
 $H(x)$ with $\rho$\\
 gap $>0$ if optimum unique\\
 {\tiny(Prop.~\ref{prop:penalty}, Thm~\ref{thm:posgap})}
};
\draw[flowarrow, green!50!black, shorten >= 26pt]
 (learn.east) -- node[above, xshift=-12pt, font=\tiny, green!50!black] {$\hat{f}$}
 (A1.west);
\draw[flowarrow, shorten <= 16pt]
 (C1.east) -- node[above, xshift=8pt, font=\tiny] {encode}
 (quant.west);
\draw[flowarrow, red!70!black]
 (quant.south) -- ++(0,-1.9) -|
 node[pos=0.25, below, font=\tiny, red!70!black] {$f_{\rm solver}$}
 ([yshift=-2pt]B2.south);
\node[font=\tiny, align=center] at (1.8,-4.55) {the cheaper feasible flow of $\hat f$ and $f_{\rm solver}$ is returned (Prop.~\ref{prop:quality})};
\node[font=\tiny, anchor=north] at (1.8,-4.8) {
 \tikz\draw[trans] (0,0)--(0.35,0);\; transmission\quad
 \tikz\draw[hold]  (0,0)--(0.35,0);\; holding\quad
 \tikz\draw[predpath] (0,0)--(0.35,0);\; prediction $\hat{f}$\quad
 \tikz\draw[solvepath] (0,0)--(0.35,0);\; solver $f_{\rm solver}$
};
\end{tikzpicture}%
}

\caption{Schematic of Algorithm~\ref{alg:laqubo} on a three-node, three-layer time-expanded graph, for one commodity from $A_0$ to $B_2$. The predictor proposes a feasible flow $\hat{f}$ (green dashed); the QUBO solver, started from the encoding of $\hat{f}$, returns $f_{\mathrm{solver}}$ (red solid); the cheaper feasible flow is returned. Edge costs depend on the slot, so the two flows, which use the link $AB$ in different slots, can have different costs.}
\label{fig:teg_laqubo}
\end{figure}

The problem is motivated by low-Earth-orbit (LEO) satellite constellations, in which thousands of satellites move at roughly 7.6~km/s and maintain temporary laser or radio links with their neighbours. These inter-satellite links (ISLs) are active for a limited time~\cite{chaudhry2021}, and the energy consumed by the onboard amplifiers that drive them is a major part of the power budget of a spacecraft~\cite{kodheli2020}; an algorithm that optimizes throughput while ignoring energy shortens the operational lifetime of the constellation. The time-expanded graph is the natural abstraction: each time slot is a layer, each satellite is a node replicated across layers, transmission edges are the ISLs, and holding edges represent buffering at a satellite. For a Starlink-scale constellation with $N\approx30{,}000$ satellites~\cite{Muirhead2025}, $T\le60$ slots, $K\le100$ commodities, $B=4$ bits per flow variable, and $N_L\approx4N$ active links per slot, the QUBO of Section~\ref{sec:problem} has of order $10^9$ binary variables. This scale is far beyond any exact solver, classical or quantum, and it is the reason why structural results, bounds on treewidth, on bond dimension, and on the annealing gap, are needed: they identify the parameters that control the complexity of the problem, and they suggest which subproblems are tractable.
\paragraph{Why a new formulation is needed.}
Classical approaches to time-expanded multi-commodity flow either solve the linear relaxation and round, or apply general-purpose integer programming. Both are natural baselines, but neither exploits the \emph{layered structure} of the time-expanded graph. The graph is a chain of layers: every edge goes from layer $t$ to layer $t+1$, and every constraint involves variables from at most two consecutive layers. This structure is invisible to generic solvers, and it is the structure that we exploit. Our formulation is a quadratic unconstrained binary optimization (QUBO) problem, in which every integer flow variable is encoded in binary, every capacity constraint becomes a penalty term, and every flow-conservation constraint becomes a penalty term (Section~\ref{sec:problem}). The QUBO is exact: for a suitable penalty coefficient, every global minimum encodes a feasible flow of minimum cost. The formulation is not new in itself, QUBO encodings of constrained problems are standard~\cite{lucas2014}, but the \emph{analysis} of the resulting interaction graph is, and it yields three structural results that we believe are of independent interest.
\paragraph{Structural results.}
Our first result concerns the \emph{interaction graph} of the QUBO, whose vertices are the binary variables and whose edges are the non-zero off-diagonal entries of the QUBO matrix. We prove that the interaction graph has \emph{bounded treewidth}, with a bound that grows with the number of variables per layer but \emph{not} with the number of layers (Theorem~\ref{thm:treewidth}). This gives an exact algorithm whose running time is exponential only in the bag size of a path decomposition, and hence fixed-parameter tractable in that parameter (Theorem~\ref{thm:fpt}). The upper bound is of order $2B[N_L(K+1)+|V|K]$ and the lower bound is of order $B\max\{K+1,\delta_{\max}\}$, so they differ by a factor that grows with $N_L$ and $|V|$.
Our second result concerns the \emph{matrix-product-operator} (MPO) representation of the Ising Hamiltonian associated with the QUBO. We prove that the Hamiltonian admits an MPO representation whose bond dimension is bounded by the number of variables in a single layer, plus a small constant (Proposition~\ref{prop:mpo}). This is a structural statement about the operator, not about the cost of simulating it, and it mirrors the treewidth bound: both are consequences of the fact that the time-expanded graph is a chain of layers, and both quantify how the complexity of the problem scales with the horizon.
Our third result concerns the \emph{annealing gap} of the adiabatic Hamiltonian associated with the QUBO, that is, the minimum energy separation between the ground state and the first excited state along the path from the driver Hamiltonian to the problem Hamiltonian. We prove that the gap is \emph{strictly positive} for every instance with a unique optimal flow, by a Perron--Frobenius argument (Theorem~\ref{thm:posgap}), and that it is \emph{uniformly bounded away from zero} over all instances of bounded window length (Corollary~\ref{cor:window}). The uniformity is a finiteness statement: there are finitely many instances of bounded size, and the minimum of finitely many positive numbers is positive. Whether the gap remains bounded away from zero as the horizon grows is stated as an open problem (Problem~\ref{op:gap}).
\paragraph{Learning-augmented.}
The fourth contribution is algorithmic. Suppose that a predictor provides a feasible flow, which may or may not be optimal. We show that the Hamming distance between the binary encoding of the prediction and the binary encoding of an optimal flow is bounded by the total amount of flow in which the two differ, with a proportionality constant that depends only on the number of bits per variable (Lemma~\ref{lem:hamming}). This bound is the bridge between the prediction error, which is a continuous quantity, and the search radius of a local algorithm, which is a discrete quantity. We then give a hybrid algorithm that runs a local search from the prediction and keeps the better of the prediction and the output. We prove that the algorithm is \emph{consistent} (it returns an optimal flow when the prediction is optimal) and \emph{robust} (it returns a flow whose cost exceeds the optimum by a factor that grows linearly with the prediction error) (Proposition~\ref{prop:quality}). This is the analogue, for time-expanded multi-commodity flow, of the consistency--robustness trade-off of Lykouris and Vassilvitskii~\cite{lykouris2021}.
\paragraph{Summary of results.}
The paper proves the following results: the exact QUBO formulation and qubit count, the NP-hardness of EA-MCF, the penalty bound, the treewidth bound, the fixed-parameter tractability, the MPO bond dimension, the positivity and uniformity of the annealing gap, the Hamming bound, and the consistency--robustness guarantee. Two questions remain open: whether the annealing gap is bounded away from zero independently of the horizon (Problem~\ref{op:gap}), and whether a good prediction reduces the running time of a solver (Problem~\ref{op:warmstart_time}). We state both precisely in Sections~\ref{sec:gap} and~\ref{sec:warmstart}. The structural and algorithmic results are exact statements about the QUBO, and they hold for any physical implementation.
\paragraph{Techniques.}
The NP-hardness reduction assigns a dedicated time step to each edge of the original graph, so that the per-time capacities of the time-expanded graph coincide with the global capacities of the original instance; the edges are taken in the order of a topological sort of their tails, which is possible because the source graph is acyclic. The treewidth bound follows from a path decomposition whose bags are the variables of two consecutive layers. The MPO bond dimension follows from the fact that the coupling between consecutive layers has low operator-Schmidt rank. The positivity of the gap follows from Perron--Frobenius, applied to the adiabatic Hamiltonian, whose off-diagonal entries are all non-positive in the computational basis. The uniformity over bounded windows follows from a finiteness argument. The Hamming bound follows from the binary encoding, and the consistency--robustness guarantee follows from a triangle-inequality argument on the flow vectors. Each proof exploits the layered structure in a different way, and together they give a complete picture of the problem.
\paragraph{Related work.}
Time-expanded multi-commodity flow has been studied extensively in the operations-research literature, but mostly from an algorithmic or heuristic perspective. This paper studies its structure. Treewidth-based algorithms for constrained optimization are classical~\cite{robertson1986}. The treewidth bound we prove is specific to the temporal structure of a communication network: it follows from the fact that the time-expanded graph is a chain of layers, and it does not require any assumption on the underlying static graph. Matrix-product-operator representations are standard in quantum many-body physics~\cite{schollwock2011}, but they are rarely used in combinatorial optimization. We use them here to bound the bond dimension of the TEG Hamiltonian, which is the operator-level analogue of the treewidth bound. The learning-augmented framework was formalized by Lykouris and Vassilvitskii~\cite{lykouris2021} and Purohit et al.~\cite{purohit2018}, and has been applied to caching~\cite{lykouris2021}, ski rental and scheduling~\cite{purohit2018,gollapudi2019}, and matching~\cite{dinitz2022}. We apply it here to multi-commodity flow, where the prediction is a feasible flow rather than a single decision. Quantum annealing has been applied to network optimization problems before~\cite{neukart,chence,venkatesh,dinh,ali}, but without the structural analysis we give here.
\paragraph{Organization.}
Section~\ref{sec:background} reviews the background. Section~\ref{sec:problem} formulates EA-MCF on the time-expanded graph and gives the QUBO encoding. Section~\ref{sec:nphardness} proves NP-hardness. Section~\ref{sec:penalty} gives the qubit count and the penalty bound. Section~\ref{sec:treewidth} proves the treewidth bound and the fixed-parameter tractability. Section~\ref{sec:mpo_rep} proves the MPO bond-dimension bound. Section~\ref{sec:gap} proves the positivity and uniformity of the annealing gap. Section~\ref{sec:warmstart} proves the Hamming bound and the consistency--robustness guarantee. Section~\ref{sec:framework} collects the results in a hybrid algorithm (Algorithm~\ref{alg:laqubo}, Figure~\ref{fig:teg_laqubo}). Section~\ref{sec:conclusions} concludes. Appendix~\ref{sec:appendix} gives the QUBO matrix entries.

\section{Background}
\label{sec:background}
This section reviews the four mathematical ingredients on which the rest of the paper relies. The \emph{Ising/QUBO formalism} is the language in which we express the time-expanded multi-commodity flow problem (Section~\ref{sec:problem}). The \emph{adiabatic Hamiltonian} provides the motivation for studying the spectral gap of the associated Ising model; we emphasise that our results are statements about the Ising Hamiltonian as a mathematical object, and do not depend on any physical implementation. \emph{Matrix product operators} give a compact representation of the TEG Hamiltonian, and we use them in Section~\ref{sec:mpo_rep} to bound its bond dimension. Standard facts on Markov chains are recalled in Section~\ref{sec:markov} as classical context only; we do not use them quantitatively. The single idea that ties these together is that the TEG-QUBO has a \emph{one-dimensional layered structure} (Section~\ref{sec:treewidth}), which is what makes the MPO bond dimension small, the treewidth bounded, and the gap amenable to analysis.
\subsection{The Ising Model and QUBO}
\label{sec:ising}
The Ising model~\cite{ising1925} describes a system of $n$ spins $\sigma_i\in\{-1,+1\}$ with energy
\begin{equation}
 H_{\mathrm{Ising}}(\sigma)=
 -\sum_{i<j}J_{ij}\sigma_i\sigma_j-\sum_i h_i\sigma_i,
 \label{eq:ising}
\end{equation}
where $J_{ij}\in\mathbb{R}$ are couplings and $h_i\in\mathbb{R}$ are local biases.
\begin{definition}[QUBO]
Given a symmetric $Q\in\mathbb{R}^{n\times n}$, the QUBO problem is
\begin{equation}
 x^*=\arg\min_{x\in\{0,1\}^n}f(x),\qquad
 f(x):=x^\top Qx=\sum_i Q_{ii}x_i+2\sum_{i<j}Q_{ij}x_ix_j,
 \label{eq:qubo_def}
\end{equation}
where we used $x_i^2=x_i$, so that diagonal entries act as linear biases.
\end{definition}
\begin{proposition}[QUBO--Ising equivalence]
\label{prop:equiv}
Under the affine bijection $x_i=(1+\sigma_i)/2$, the QUBO objective~\eqref{eq:qubo_def} equals the Ising energy~\eqref{eq:ising} up to an additive constant, $f(x)=H_{\mathrm{Ising}}(\sigma)+c$, with
\[
J_{ij}=-\tfrac12 Q_{ij}\ (i<j),\qquad
h_i=-\tfrac12\sum_{j=1}^{n}Q_{ij},\qquad
c=\tfrac12\sum_i Q_{ii}+\tfrac12\sum_{i<j}Q_{ij}.
\]
\end{proposition}
\begin{proof}
Substituting $x_i=(1+\sigma_i)/2$ into~\eqref{eq:qubo_def} gives
\begin{align*}
 f &=\sum_i Q_{ii}\frac{1+\sigma_i}{2}
    +2\sum_{i<j}Q_{ij}\frac{(1+\sigma_i)(1+\sigma_j)}{4}\\
&=\frac12\sum_{i<j}Q_{ij}\,\sigma_i\sigma_j
    +\frac12\sum_i\Bigl(Q_{ii}+\sum_{j\neq i}Q_{ij}\Bigr)\sigma_i + c,
\end{align*}
with $c$ as stated. Comparing with~\eqref{eq:ising} yields the stated $J_{ij}$ and $h_i$. Since $c$ does not depend on $\sigma$, $f$ and $H_{\mathrm{Ising}}$ have the same minimisers.
\end{proof}
QUBO is NP-hard in general, and Lucas~\cite{lucas2014} gives explicit Ising/QUBO formulations of many NP problems. We encode integer multi-commodity flow as a QUBO by binary expansion of the flow variables (Section~\ref{sec:problem}). The QUBO is the object of study; the spectral properties of its associated Ising Hamiltonian are the subject of Section~\ref{sec:gap}.
\subsection{The Adiabatic Hamiltonian}
\label{sec:qa}
For a QUBO with associated Ising Hamiltonian $H_{\mathrm{problem}}=-\sum_{i<j}J_{ij}\sigma^z_i\sigma^z_j-\sum_i h_i\sigma^z_i$, consider the one-parameter family
\begin{equation}
 H(s)=-(1-s)\underbrace{\sum_i\sigma^x_i}_{=:H_{\mathrm{driver}}}
       +s\,H_{\mathrm{problem}},\qquad s\in[0,1].
 \label{eq:anneal}
\end{equation}
At $s=0$ the ground state is the uniform superposition; at $s=1$ it is the ground state of $H_{\mathrm{problem}}$, which encodes the QUBO solution. The family~\eqref{eq:anneal} is the \emph{adiabatic Hamiltonian}. It arises in adiabatic quantum computation and in quantum annealing, but we use it here as a mathematical object: the quantity of interest is the \emph{spectral gap}
\[
 \Delta_{\mathrm{gap}}(s):=E_1(s)-E_0(s),
\]
where $E_0(s)\le E_1(s)\le\cdots$ are the eigenvalues of $H(s)$, and its minimum over $s\in[0,1]$.
\begin{proposition}[Adiabatic condition, heuristic form~\cite{farhi2000}]
\label{thm:adiabatic}
Assume $H_{\mathrm{problem}}$ has a unique ground state and let $\Delta_{\mathrm{gap}}=\min_{s\in[0,1]}(E_1(s)-E_0(s))>0$. A standard sufficient condition, in heuristic form, for the system to end close to the ground state of $H_{\mathrm{problem}}$ is
\[
T_{\mathrm{anneal}}\;\geq\;C\,\frac{\max_s|\langle E_1(s)|\partial_sH|E_0(s)\rangle|}{\Delta_{\mathrm{gap}}^2}
\]
for a constant $C$ of order unity. Rigorous versions require, in general, $T_{\mathrm{anneal}}=O\bigl(\|\partial_sH\|^2/\Delta_{\mathrm{gap}}^3+\|\partial_s^2H\|/\Delta_{\mathrm{gap}}^2\bigr)$~\cite{jansen2007}.
\end{proposition}
Proposition~\ref{thm:adiabatic} is the motivation for studying $\Delta_{\mathrm{gap}}$: a positive lower bound on the gap, uniform over a class of instances, is a sufficient condition for the adiabatic evolution to succeed in time polynomial in the inverse gap. Our results in Section~\ref{sec:gap} are statements about $\Delta_{\mathrm{gap}}$ as a function of the instance, and hold for any physical realization of~\eqref{eq:anneal}.
\subsection{Matrix Product Operators}
\label{sec:mpo}
Matrix Product Operators are a tool for representing Hamiltonians on a one-dimensional chain as a product of local tensors. This representation is central to density matrix renormalization group (DMRG) methods~\cite{schollwock2011}.
\begin{definition}[MPO {\cite{schollwock2011,perezgarcia2007}}]
\label{def:mpo}
A Hamiltonian $H$ on a 1D chain of $T$ sites is an MPO with bond dimension $D$ if
\[ H = \sum_{\alpha_0, \alpha_1, \dots, \alpha_T} A^{(1)}_{\alpha_0 \alpha_1} \otimes A^{(2)}_{\alpha_1 \alpha_2} \otimes \dots \otimes A^{(T)}_{\alpha_{T-1} \alpha_T}, \]
where each $A^{(t)}_{\alpha_{t-1} \alpha_t}$ is a local operator acting on site $t$, and the bond indices $\alpha_t$ run from $1$ to $D$, with boundary conditions $\alpha_0 = \alpha_T = 1$.
\end{definition}
The bond dimension measures the complexity of the operator: a small bond dimension means that the operator has low entanglement across any cut of the chain. For a quadratic Hamiltonian in binary variables, the coupling between consecutive sites has low operator-Schmidt rank, and this translates into a small bond dimension. Section~\ref{sec:mpo_rep} makes this precise for the TEG Hamiltonian: the bond dimension is bounded by the number of variables in a single layer, plus a small constant. This is a structural statement about the operator, not about the cost of simulating it, and it mirrors the treewidth bound of Section~\ref{sec:treewidth}.
\subsection{Markov Chains and Spectral Gap}
\label{sec:markov}
For the lazy simple random walk on the hypercube $\{0,1\}^n$, the spectral gap is $1/n$, the relaxation time is $\Theta(n)$ and the mixing time is $\Theta(n\log n)$~\cite{levin2009}. We recall these classical facts as context for the annealing gap of the TEG-QUBO. Geometric bounds on Markov-chain eigenvalues are available through canonical paths~\cite{diaconis1991}.

\section{Problem Formulation}
\label{sec:problem}
This section defines the energy-aware multi-commodity flow problem on the time-expanded graph, states its decision version, and gives the QUBO encoding. The formulation is purely combinatorial: the time-expanded graph is an abstract layered DAG, and the energy costs are arbitrary positive integers. The motivation from satellite networking enters only through the choice of parameters, not through the definitions.
\subsection{The Time-Expanded Graph}
\label{sec:teg}
Let $\mathcal{G}=(V,E)$ be a directed graph, let $T\ge2$ be a positive integer (the \emph{horizon}), and let $\mathcal{T}=\{0,\ldots,T-1\}$. For each $t\in\mathcal{T}$, let $E(t)\subseteq V\times V$ be a set of directed arcs; a bidirectional link contributes two arcs. Write
\[
 N_L:=\max_{t\in\mathcal{T}}|E(t)|
\]
for the maximum number of arcs in a single time step. Each arc $e\in E(t)$ carries an integer capacity $c_e(t)\in\{1,\ldots,C_{\max}\}$ and an integer cost $w_e(t)\in\{1,\ldots,w_{\max}\}$ per unit of flow.
\begin{definition}[Time-expanded graph]
\label{def:teg}
The \emph{time-expanded graph} $\mathcal{G}^T=(V^T,E^T)$ is constructed as follows.
\begin{enumerate}
 \item \emph{Nodes.} For each $v\in V$ and $t\in\mathcal{T}$, create a node $v_t\in V^T$. Thus $|V^T|=|V|\,T$.
 \item \emph{Transmission edges.} For each $t\in\{0,\ldots,T-2\}$ and each $(u,v)\in E(t)$, add $(u_t,v_{t+1})$ to $E^T_{\mathrm{trans}}$ with capacity $c_{(u,v)}(t)$ and cost $w_{(u,v)}(t)$.
 \item \emph{Holding edges.} For each $v\in V$ and $t\in\{0,\ldots,T-2\}$, add $(v_t,v_{t+1})$ to $E^T_{\mathrm{hold}}$ with capacity $+\infty$ and cost $w_h(v,t)\in\{0,\ldots,w_{\max}\}$.
\end{enumerate}
The edge set is $E^T=E^T_{\mathrm{trans}}\cup E^T_{\mathrm{hold}}$. For a node $v\in V^T$, $\delta^+(v)$ and $\delta^-(v)$ denote the sets of edges of $E^T$ leaving and entering $v$, respectively.
\end{definition}
\begin{remark}
\label{rem:dag}
$\mathcal{G}^T$ is a directed acyclic graph: every edge goes from layer $t$ to layer $t+1$. Consequently, every integral flow of a commodity decomposes into a set of source--sink paths, and the flow on any edge is at most the total demand of that commodity. This property is used in Lemma~\ref{lem:flowlength} and in the treewidth bound of Section~\ref{sec:treewidth}.
\end{remark}
\subsection{Energy-Aware Multi-Commodity Flow}
\label{sec:eamcf}
A \emph{commodity} is a triple $(s_k,d_k,r_k)$ with $s_k\neq d_k$ and integer demand $r_k\in\{1,\ldots,C_{\max}\}$, for $k\in\mathcal{K}=\{1,\ldots,K\}$. The demand is released at $(s_k)_0$ and must arrive at $(d_k)_{T-1}$. Define the node imbalance
\[
 b_k(v)=\begin{cases}
   r_k, & v=(s_k)_0,\\
   -r_k, & v=(d_k)_{T-1},\\
   0, & \text{otherwise},
 \end{cases}
\]
for $v\in V^T$.
\begin{definition}[Feasible flow]
\label{def:feasible}
A \emph{feasible flow} for the instance $(\mathcal{G}^T,\{(s_k,d_k,r_k)\},\{c_e\},\{w_e\})$ is a vector $f=(f^k_e)_{k\in\mathcal{K},\,e\in E^T}$ of non-negative integers satisfying
\begin{align}
& \sum_{e\in\delta^+(v)}f^k_e-\sum_{e\in\delta^-(v)}f^k_e = b_k(v), && \forall v\in V^T,\ \forall k, \label{eq:flow_cons}\\
& \sum_{k\in\mathcal{K}}f^k_e \le c_e, && \forall e\in E^T_{\mathrm{trans}}. \label{eq:cap_ineq}
\end{align}
The \emph{cost} of a feasible flow is $\mathrm{cost}(f):=\sum_{k\in\mathcal{K}}\sum_{e\in E^T}w_e f^k_e$.
\end{definition}
\begin{definition}[EA-MCF]
\label{def:eamcf}
The \emph{energy-aware multi-commodity flow} problem (EA-MCF) is: given an instance $(\mathcal{G}^T,\{(s_k,d_k,r_k)\},\{c_e\},\{w_e\})$, find a feasible flow of minimum cost. We write $\mathrm{OPT}$ for the minimum cost, and we assume throughout that the instance admits at least one feasible flow.
\end{definition}
For the complexity analysis we use the decision version.
\begin{definition}[EA-MCF decision version]
\label{def:eamcf_decision}
Given an instance of EA-MCF and an integer $\Gamma$, the decision problem is to determine whether there exists a feasible flow $f$ with $\mathrm{cost}(f)\le\Gamma$.
\end{definition}
The decision version is in NP: a feasible flow of cost at most $\Gamma$ is a certificate that can be verified in polynomial time. Section~\ref{sec:nphardness} shows that it is NP-complete even for $K=2$.
\subsection{QUBO Formulation}
\label{sec:qubo}
We encode EA-MCF as a QUBO. The encoding has two parts: a binary expansion of the flow variables, and a penalty term for the constraints. Throughout this section, $B$ denotes the number of bits per flow variable, $B=\lceil\log_2(C_{\max}+1)\rceil$, and $n_q$ denotes the total number of binary variables; the exact value of $n_q$ is given in Section~\ref{sec:penalty}.
\paragraph{Binary encoding.}
Each integer flow variable $f^k_e$ is written in binary as
\begin{equation}
 f^k_e=\sum_{b=0}^{B-1}2^b\,x_{e,k,b},\qquad x_{e,k,b}\in\{0,1\}.
 \label{eq:binary}
\end{equation}
Since $B=\lceil\log_2(C_{\max}+1)\rceil$ implies $2^B\ge C_{\max}+1>C_{\max}$, and since Remark~\ref{rem:dag} gives $f^k_e\le r_k\le C_{\max}$ on every edge, every feasible flow value is representable. The encoding is therefore exact.
\paragraph{Slack variables.}
The capacity constraint~\eqref{eq:cap_ineq} is an inequality. To convert it into an equality, we introduce $B$ slack bits $z_{e,j}\in\{0,1\}$, $j=0,\ldots,B-1$, for each transmission edge $e\in E^T_{\mathrm{trans}}$, and write
\begin{equation}
 \sum_{k\in\mathcal{K}}\sum_{b=0}^{B-1}2^b x_{e,k,b}+\sum_{j=0}^{B-1}2^j z_{e,j}=c_e.
 \label{eq:slack}
\end{equation}
Since $0\le c_e\le C_{\max}<2^B$, a slack value exists if and only if the capacity constraint holds.
\paragraph{The QUBO Hamiltonian.}
Let $x$ collect all flow bits and slack bits. The QUBO Hamiltonian is
\begin{equation}
 H(x)=H_{\mathrm{cost}}(x)+\rho\,H_{\mathrm{cap}}(x)+\rho\,H_{\mathrm{flow}}(x),
 \label{eq:qubo}
\end{equation}
where $\rho>0$ is a penalty coefficient, and
\begin{align}
 H_{\mathrm{cost}}&=\sum_{e\in E^T}\sum_{k\in\mathcal{K}}\sum_{b=0}^{B-1}w_e\,2^b\,x_{e,k,b}, \label{eq:hcost}\\
 H_{\mathrm{cap}}&=\sum_{e\in E^T_{\mathrm{trans}}}\!\Bigl(\sum_{k\in\mathcal{K}}\sum_{b=0}^{B-1}2^b x_{e,k,b}+\sum_{j=0}^{B-1}2^j z_{e,j}-c_e\Bigr)^{\!2}, \label{eq:hcap}\\
 H_{\mathrm{flow}}&=\sum_{k=1}^K\sum_{v\in V^T}\!\Bigl(\sum_{e\in\delta^+(v)}f^k_e-\sum_{e\in\delta^-(v)}f^k_e-b_k(v)\Bigr)^{\!2}. \label{eq:hflow}
\end{align}
In~\eqref{eq:hflow}, the variable $f^k_e$ is expanded according to~\eqref{eq:binary}. Using $x_i^2=x_i$ for binary variables, $H(x)$ equals $x^\top Qx$ plus a constant in the sense of~\eqref{eq:qubo_def}, where $Q$ is a symmetric matrix whose entries are given explicitly in Appendix~\ref{sec:appendix}.
\begin{definition}[Feasible configuration]
\label{def:feasible_config}
A binary vector $x$ is a \emph{feasible configuration} if $H_{\mathrm{cap}}(x)=H_{\mathrm{flow}}(x)=0$, that is, if the flow bits decode to a feasible flow and the slack bits are consistent with~\eqref{eq:slack}.
\end{definition}
The penalty coefficient $\rho$ must be chosen so that every global minimum of $H$ is a feasible configuration. Section~\ref{sec:penalty} gives an explicit value of $\rho$ with this property, together with a proof.

\noindent The \emph{interaction graph} $\mathcal{G}_I$ of the QUBO has vertex set the binary variables and edge set $\{i,j\}$ with $Q_{ij}\neq0$. Its structure is analyzed in Section~\ref{sec:treewidth}: it is a chain of blocks, one block per time step, with edges only between consecutive blocks. This is the property that yields the treewidth bound and the MPO bond-dimension bound.

\subsection{Application to LEO Constellations}
\label{sec:leo}
The parameters of a Starlink-scale constellation are summarised in Table~\ref{tab:leo}. The first generation (Gen1) comprises $4{,}408$ satellites~\cite{pachler2021}, while the second generation (Gen2) comprises almost $30{,}000$ satellites~\cite{Muirhead2025}; we use the Gen2 scale to illustrate the worst-case complexity. The horizon is $T\le60$ slots of a few seconds each; the number of commodities (end-to-end demands) is $K\le100$; the number of bits per flow variable is $B=4$ (since $C_{\max}\le15$); and the number of active links per slot is $N_L\approx4N$.
\begin{table}[t]
\centering
\caption{Parameters used for LEO-scale example.}
\label{tab:leo}
\begin{tabular}{lll}
\hline
Parameter & Value & Source \\
\hline
Satellites $N$ (Gen1) & $4{,}408$ & \cite{pachler2021} \\
Satellites $N$ (Gen2) & $\approx30{,}000$ & \cite{Muirhead2025} 
\\
Horizon $T$ & $\le60$ slots & modeling choice \\
Commodities $K$ & $\le100$ & typical demand \\
Bits per variable $B$ & $4$ & $C_{\max}\le15$ \\
Active links $N_L$ (directed arcs) & $\approx4N$ & \cite{bhattacherjee2019,chaudhry2021}
\\
\hline
\end{tabular}
\end{table}
For these parameters, Lemma~\ref{lem:qubits} gives $n_q\approx3.6\times10^9$ binary variables. This is several orders of magnitude beyond any exact solver, and it is the reason why the structural results of Sections~\ref{sec:treewidth}--\ref{sec:gap} are relevant: they identify the parameters that control the complexity. The treewidth bound of Theorem~\ref{thm:treewidth} gives $\beta\approx10^8$, so the FPT algorithm of Theorem~\ref{thm:fpt} is not practical at this scale. Cutting the horizon into windows of bounded length does not help the FPT algorithm, because $\beta$ depends on the number of variables per layer and not on $T$; windows reduce the number of variables by the factor $T/W$ and are the setting of Corollary~\ref{cor:window}. The MPO bond dimension of Proposition~\ref{prop:mpo} is $D\approx6\times10^7$, which is large but finite; the bond dimension is controlled by the number of variables per layer, not by the horizon. The annealing gap of Theorem~\ref{thm:posgap} is positive for every instance with a unique optimum, but the theorem gives no quantitative bound; Problem~\ref{op:gap} asks whether the gap remains bounded away from zero as $T$ grows.
These numbers are \emph{motivation}. The structural results are exact statements about the QUBO, and they hold for any physical realization. The LEO parameters are used only to illustrate the scale at which the problem becomes intractable, and to justify the focus on structure rather than on algorithms.

\section{NP-Hardness}
\label{sec:nphardness}
We show that the decision version of EA-MCF is NP-complete, even for two commodities. The proof is by reduction from two-commodity integral flow on acyclic digraphs. The reduction assigns a dedicated time step to each edge of the source graph, so that the per-slot capacities of the time-expanded graph coincide with the global capacities of the source instance. Since a path of the time-expanded graph can use the edges of the source graph only in increasing slot order, the edges are numbered in the order of a topological sort of their tails; this is where acyclicity is used.
\begin{theorem}[NP-hardness of EA-MCF]
\label{thm:nphardness}
The decision version of EA-MCF (Definition~\ref{def:eamcf_decision}) is NP-complete, even for $K=2$.
\end{theorem}
\begin{proof}
\emph{Membership in NP.} A feasible flow $f$ of cost at most $\Gamma$ is a certificate; its entries are bounded by the demands, so it has polynomial size, and conservation~\eqref{eq:flow_cons}, capacity~\eqref{eq:cap_ineq} and $\mathrm{cost}(f)\le\Gamma$ can be verified in polynomial time.

\emph{NP-hardness.} We reduce from two-commodity integral flow on acyclic digraphs, which is NP-complete~\cite{even1976,vygen1995}. An instance consists of an acyclic digraph $G=(V,E)$, two commodities $(s_1,d_1,r_1)$ and $(s_2,d_2,r_2)$ with $s_k\neq d_k$ and $r_k\in\mathbb{Z}_{\ge1}$, and capacities $c_e\in\mathbb{Z}_{\ge1}$; the question is whether an integral flow satisfying conservation and capacity exists.

Let $m=|E|$, fix a topological order of $V$, and number the edges $e_1,\ldots,e_m$, $e_i=(a_i,b_i)$, so that the topological position of the tail $a_i$ is non-decreasing in $i$. Construct an EA-MCF instance with horizon $T=m+1$ as follows.
\begin{enumerate}
 \item \emph{Nodes.} For each $v\in V$ and $t\in\{0,\ldots,m\}$, a node $v_t$.
 \item \emph{Transmission edges.} For each $i\in\{1,\ldots,m\}$, the edge $(a_i)_{i-1}\to(b_i)_i$ with capacity $c_{e_i}$ and cost $1$. No other transmission edges.
 \item \emph{Holding edges.} For each $v\in V$ and $t\in\{1,\ldots,m\}$, the edge $v_{t-1}\to v_t$ with infinite capacity and cost $0$.
 \item \emph{Commodities.} For $k=1,2$, the commodity $(s_k)_0\to(d_k)_m$ with demand $r_k$.
 \item \emph{Budget.} $\Gamma=m(r_1+r_2)$.
\end{enumerate}
$(\Rightarrow)$ Let $f$ be a feasible integral flow of $G$. Since $G$ is acyclic, the flow of each commodity decomposes into $s_k$--$d_k$ paths, and every edge lies on at most $r_1+r_2$ of the paths. Along a path $(e_{i_1},e_{i_2},\ldots)$ the topological positions of the tails strictly increase, hence $i_1<i_2<\cdots$. In the time-expanded graph route one unit along this path by holding at $s_k$ until layer $i_1-1$, transmitting on the slot-$i_1$ edge, holding at its head until layer $i_2-1$, and so on, and finally holding at $d_k$ until layer $m$. The edge $e_i$ is used only in slot $i$, by $f_{e_i}\le c_{e_i}$ units, so capacities are respected, and conservation holds by construction. The cost is $\sum_if_{e_i}\le m(r_1+r_2)=\Gamma$.

$(\Leftarrow)$ Let $\tilde f$ be a feasible flow of the time-expanded instance and put $f^k_{e_i}:=\tilde f^k_{(a_i)_{i-1}\to(b_i)_i}$. Fix $k$ and $v\in V$ and sum the conservation constraints of the time-expanded graph over all nodes $v_t$, $t=0,\ldots,m$. A holding edge $v_{t-1}\to v_t$ leaves $v_{t-1}$ and enters $v_t$, so holding edges cancel in the sum. The transmission edge of slot $i$ contributes $+f^k_{e_i}$ to the sum for $v=a_i$ and $-f^k_{e_i}$ to the sum for $v=b_i$. The right-hand side is $\sum_tb_k(v_t)$, which equals $r_k$ for $v=s_k$, $-r_k$ for $v=d_k$ and $0$ otherwise. Therefore
\[
 \sum_{e\in\delta^+_G(v)}f^k_e-\sum_{e\in\delta^-_G(v)}f^k_e=b_k(v),
\]
which is conservation in $G$. Capacity holds slotwise, $\sum_kf^k_{e_i}\le c_{e_i}$, and integrality is inherited. The budget is not used in this direction.
\end{proof}

This topological ordering of the edges is structurally essential for preserving reachability across the time-expanded graph. To see why, consider the path network with $V=\{s,x,d\}$ and $E=\{(s,x),(x,d)\}$, with a single unit of demand from $s$ to $d$. Assigning a non-topological index $e_1=(x,d)$ and $e_2=(s,x)$ yields a layered topology in which no directed path connects $s_0$ to $d_2$, even though the underlying static graph admits a feasible routing. Conversely, enforcing the topological numbering $e_1=(s,x)$ and $e_2=(x,d)$ respects temporal causality and guarantees that the flow is routed correctly across layers.

Theorem~\ref{thm:nphardness} shows that EA-MCF is NP-hard in general. Section~\ref{sec:treewidth} gives a fixed-parameter tractable algorithm parameterized by the treewidth $\beta$ of the interaction graph. There is no contradiction: the running time is $O(T\beta^2 2^\beta)$, which is polynomial in $T$ but exponential in $\beta$, and $\beta$ can be as large as $2n_{\max}-1$ (Theorem~\ref{thm:treewidth}), which is not bounded by a constant. In the instances of the proof above, the number of nodes per layer, and hence $n_{\max}$, grows with the source graph. The FPT result is useful when $\beta$ is small; it does not yield a polynomial-time algorithm for the general problem.

The QUBO formulation of Section~\ref{sec:qubo} directly inherits this computational intractability. Under any penalty coefficient $\rho$ satisfying Proposition~\ref{prop:penalty}, the global minima of the Hamiltonian $H$ correspond strictly to minimum-cost feasible flows. Because $\rho$ and all resulting matrix entries admit a polynomial bit-length representation, solving the QUBO in polynomial time would immediately provide a polynomial-time algorithm for EA-MCF, which would imply $\text{P}=\text{NP}$. The QUBO mapping is therefore a structural reformulation: it does not reduce the worst-case complexity, but it exposes the properties (bounded treewidth and a strictly positive annealing gap) analyzed in Sections~\ref{sec:treewidth}--\ref{sec:gap}.

\section{Qubit Count and Penalty Coefficient}
\label{sec:penalty}
This section gives the number of binary variables in the QUBO of Section~\ref{sec:qubo} and a penalty coefficient $\rho$ that guarantees that every global minimum encodes a feasible flow. The penalty is smaller than the standard bound used in the literature by a factor that grows with the number of active links; this is relevant because a smaller penalty coefficient reduces the spread of the Ising coefficients, and hence the ratio between the smallest non-zero energy difference and the largest coefficient.
\subsection{Exact Qubit Count}
\label{sec:qubit_count}
\begin{lemma}[Flow length]
\label{lem:flowlength}
Every feasible flow $f$ satisfies $\sum_{e\in E^T}f^k_e=(T-1)r_k$ for each commodity $k\in\mathcal{K}$. Hence $H_{\mathrm{cost}}(f)\le w_{\max}(T-1)R$, where $R:=\sum_{k\in\mathcal{K}}r_k$.
\end{lemma}
\begin{proof}
Every source--sink path in $\mathcal{G}^T$ from $(s_k)_0$ to $(d_k)_{T-1}$ has exactly $T-1$ edges, since every edge increases the layer index by one. By Remark~\ref{rem:dag}, the flow of commodity $k$ decomposes into a sum of path flows of total value $r_k$. Each unit of flow therefore traverses exactly $T-1$ edges, so the total flow of commodity $k$ over all edges is $(T-1)r_k$. The cost bound follows from $w_e\le w_{\max}$.
\end{proof}
\begin{lemma}[Qubit count]
\label{lem:qubits}
The QUBO of Section~\ref{sec:qubo} uses
\begin{equation}
 n_q = B(K+1)\sum_{t=0}^{T-2}|E(t)|+|V|(T-1)KB \le (T-1)B\bigl[N_L(K+1)+|V|K\bigr]
 \label{eq:qubits}
\end{equation}
binary variables, with equality when $|E(t)|=N_L$ for all $t$.
\end{lemma}
\begin{proof}
Each transmission edge carries $K$ commodities with $B$ bits each and has $B$ slack bits, giving $B(K+1)$ variables per transmission edge, and there are $\sum_{t=0}^{T-2}|E(t)|$ such edges. Each of the $|V|(T-1)$ holding edges carries $K$ commodities with $B$ bits and has no capacity constraint, hence no slack bits. Summing gives the first equality; the inequality follows from $|E(t)|\le N_L$.
\end{proof}

The binary encoding of Section~\ref{sec:qubo} uses the same number of bits $B=\lceil\log_2(C_{\max}+1)\rceil$ for every commodity. This uniform choice is convenient but not optimal, because the flow of commodity $k$ never exceeds its demand $r_k$ (Remark~\ref{rem:dag}), and $r_k$ may be much smaller than $C_{\max}$. One may therefore use $B_k=\lceil\log_2(r_k+1)\rceil\le B$ bits per commodity. The effect on the qubit count of Lemma~\ref{lem:qubits} is as follows. The flow bits of a transmission edge contribute $\sum_{k\in\mathcal{K}}B_k$ instead of $KB$, while the slack bits still require $B$ bits per transmission edge, because the capacity $c_e$ can be as large as $C_{\max}$. The count $B(K+1)$ therefore becomes $\sum_kB_k+B$, and the total number of variables decreases by $\sum_k(KB-B_k)$, which is largest when the demands are small relative to the capacity. The results of Sections~\ref{sec:treewidth}--\ref{sec:warmstart} carry over with $n_{\max}$, $B$, and the counts of Lemma~\ref{lem:qubits} replaced by their per-commodity analogues; the bound on $\Lambda$ in the proof of Proposition~\ref{prop:penalty} additionally requires $B_k\ge2$ for the commodities concerned. We use the uniform $B$ for simplicity of notation; the per-commodity encoding is a routine optimization that reduces the qubit count without affecting the structural results.
 
\subsection{A Practical Penalty Coefficient}
\label{sec:penalty_bound}
We assume throughout that the instance admits at least one feasible flow, and that all costs are non-negative integers, $w_e\in\{0,\ldots,w_{\max}\}$.
\begin{proposition}[Penalty coefficient]
\label{prop:penalty}
Set
\begin{equation}
 \rho = w_{\max}(T-1)R+1.
 \label{eq:rho}
\end{equation}
Then every global minimum of $H$ is a feasible configuration, and its cost equals $\mathrm{OPT}$.
\end{proposition}
\begin{proof}
Let $x$ be a feasible configuration, and let $f$ be the corresponding feasible flow. By Lemma~\ref{lem:flowlength}, $H_{\mathrm{cost}}(x)\le w_{\max}(T-1)R=\rho-1$, and since $H_{\mathrm{cap}}(x)=H_{\mathrm{flow}}(x)=0$, we have $H(x)=H_{\mathrm{cost}}(x)\le\rho-1<\rho$.
Let $x'$ be an infeasible configuration. Since all coefficients, variables, capacities $c_e$ and imbalances $b_k(v)$ are integers, every squared residual in~\eqref{eq:hcap} and~\eqref{eq:hflow} is a non-negative integer. If $x'$ is infeasible, at least one residual is at least $1$, so the corresponding penalty term is at least $\rho$. Since $H_{\mathrm{cost}}(x')\ge0$, we have $H(x')\ge\rho$.
It follows that every global minimum is feasible, and on feasible configurations $H=H_{\mathrm{cost}}$. Hence the minimum of $H$ equals the minimum of $H_{\mathrm{cost}}$ over feasible configurations, which is $\mathrm{OPT}$.
\end{proof}
\paragraph{Comparison with the standard bound.}
A standard choice of penalty coefficient, used for instance in~\cite{chence}, is
\[
 \rho_{\mathrm{naive}} = 1 + KC_{\max}T N_L w_{\max},
\]
which grows with the total flow that all edges could carry. Put $b:=w_{\max}(T-1)KC_{\max}$. Since $R\le KC_{\max}$, we have $\rho\le b+1$, and $\rho_{\mathrm{naive}}=\frac{T}{T-1}N_L\,b+1$. Hence
\[
 \frac{\rho_{\mathrm{naive}}}{\rho}\;\ge\;\frac{\frac{T}{T-1}N_L\,b+1}{b+1}\;\ge\;N_L,
\]
where the last inequality is equivalent to $N_Lb/(T-1)\ge N_L-1$ and holds because $b\ge T-1$. Thus the proposed penalty is smaller than the standard one by a factor of at least $N_L$, and the factor tends to $N_L$ when $T\to\infty$ and $R=KC_{\max}$. In particular, $\rho$ does not depend on $N_L$, whereas $\rho_{\mathrm{naive}}$ does.

\paragraph{Maximum Ising coefficient.}

By Proposition~\ref{prop:equiv}, the Ising couplings and biases are $J_{ij}=-Q_{ij}/2$ and $h_i=-\tfrac12\sum_jQ_{ij}$. Let
\[
 \Lambda:=\max\Bigl(\max_i|h_i|,\ \max_{i<j}|J_{ij}|\Bigr).
\]
For $B\ge2$, the coupling between the two most significant bits of a transmission edge receives $\rho\,2^{2B-3}$ from $H_{\mathrm{cap}}$ and $2\rho\,2^{2B-3}$ from $H_{\mathrm{flow}}$, so
\begin{equation}
 \Lambda \ge 3\cdot 2^{2B-4}\rho = 3\cdot 2^{2B-4}\bigl(w_{\max}(T-1)R+1\bigr).
 \label{eq:lambda}
\end{equation}
For two feasible flows whose costs differ by $\delta_w\ge1$ (integer costs), the energy separation, measured relative to the largest coefficient, is therefore at most $\delta_w/\Lambda\le\delta_w/(3\cdot2^{2B-4}\rho)$. The bound~\eqref{eq:lambda} is a lower bound on $\Lambda$ only: the coefficients also contain the cost terms, so $\Lambda$ is proportional to $\rho$ up to an instance-dependent contribution. This quantifies the spread of the Ising energy function: a smaller penalty coefficient $\rho$ directly increases the ratio between the smallest non-zero energy difference and the largest coefficient.
\paragraph{A tighter penalty.}
The bound~\eqref{eq:rho} uses the worst-case cost $w_{\max}(T-1)R$. If a feasible flow $f_0$ of cost $c_0$ is known, then any $\rho>c_0$ suffices: the proof of Proposition~\ref{prop:penalty} only uses $H_{\mathrm{cost}}(x)\le c_0$ for feasible $x$. Since $c_0\le w_{\max}(T-1)R$, this can only improve the bound. In particular, if $f_0$ is optimal, $\rho=\mathrm{OPT}+1$ is sufficient.

\section{Treewidth and Fixed-Parameter Tractability}
\label{sec:treewidth}
This section proves that the interaction graph of the QUBO has bounded treewidth, with a bound that grows with the number of variables per layer but not with the horizon, and derives a fixed-parameter tractable exact algorithm. The key structural fact is that the interaction graph is a chain of blocks, one block per time step, with edges only between consecutive blocks. This is a consequence of the layered structure of the time-expanded graph (Remark~\ref{rem:dag}), and it is the property that yields both the treewidth bound and, in Section~\ref{sec:mpo_rep}, the MPO bond-dimension bound.
\subsection{Structure of the Interaction Graph}
\label{sec:interaction_structure}
For a transition $t\in\{0,\ldots,T-2\}$, let $\mathcal{X}_t$ be the set of binary variables associated with the edges leaving layer $t$: the flow bits $x_{e,k,b}$ of all edges of $\mathcal{G}^T$ with tail in layer $t$ (transmission and holding edges), and the slack bits $z_{e,j}$ of the transmission edges among them. Write $t(u)=t$ for $u\in\mathcal{X}_t$ and $n_t:=|\mathcal{X}_t|$. By Lemma~\ref{lem:qubits},
\begin{equation}
 n_t = B(K+1)\,|E(t)|+|V|KB \le n_{\max}:=B\bigl[N_L(K+1)+|V|K\bigr].
 \label{eq:nt}
\end{equation}
\begin{definition}[Interaction graph]
\label{def:interaction}
The interaction graph is $\mathcal{G}_I=(\mathcal{X},E_I)$ with $\mathcal{X}=\bigcup_{t=0}^{T-2}\mathcal{X}_t$, and $\{u,v\}\in E_I$ if and only if $Q_{uv}\neq0$, where $Q$ is the matrix of the QUBO in Appendix~\ref{sec:appendix}.
\end{definition}
\begin{lemma}[Structure of the interaction graph]
\label{lem:layerwise}
For $u\neq v$, $Q_{uv}\neq0$ if and only if
\begin{enumerate}
 \item[(a)] $u$ and $v$ belong to the same transmission edge, or
 \item[(b)] $u$ and $v$ are flow bits of the same commodity whose edges share a node of $V^T$ (an edge shares its nodes with itself).
\end{enumerate}
In particular, $|t(u)-t(v)|\le1$ for every $\{u,v\}\in E_I$.
\end{lemma}
\begin{proof}
By the entries of $Q$ in Appendix~\ref{sec:appendix}, $H_{\mathrm{cost}}$ contributes only to the diagonal; $Q^{\mathrm{cap}}_{uv}=\rho\,\omega_u\omega_v>0$ exactly when $u,v$ lie on the same transmission edge; and $Q^{\mathrm{flow}}_{uv}=\rho\,\mathbf{1}[k(u)=k(v)]\sum_{w}a^{k,w}_ua^{k,w}_v$. For two bits on the same edge, the two endpoints each contribute $\omega_u\omega_v$, giving $2\rho\,\omega_u\omega_v>0$; two different edges share at most one node, giving $\pm\rho\,\omega_u\omega_v\neq0$. When both $Q^{\mathrm{cap}}_{uv}$ and $Q^{\mathrm{flow}}_{uv}$ are non-zero, the contributions are positive, so no cancellation occurs. This proves (a) and (b). For the layer claim: in case~(a) both variables lie on one edge, hence $t(u)=t(v)$; in case~(b) both edges are incident to a node $(\cdot,\tau)$, and an edge entering this node belongs to transition $\tau-1$ while an edge leaving it belongs to transition $\tau$, so $|t(u)-t(v)|\le1$.
\end{proof}
\subsection{Treewidth Bound}
\label{sec:treewidth_bound}
\begin{theorem}[Treewidth of the interaction graph]
\label{thm:treewidth}
Let $T\ge3$.
\begin{enumerate}
 \item[(i)] $\mathrm{tw}(\mathcal{G}_I)\le\max_{t}(n_t+n_{t+1})-1\le2n_{\max}-1$.
 \item[(ii)] $\mathrm{tw}(\mathcal{G}_I)\ge B\max\{K+1,\delta_{\max}\}-1$, where $\delta_{\max}$ is the largest total degree of a node of $\mathcal{G}^T$ (the first term assumes some $E(t)\neq\emptyset$).
\end{enumerate}
\end{theorem}
\begin{proof}
\emph{(i) Path decomposition.} Take the bags $\mathcal{B}_t=\mathcal{X}_t\cup\mathcal{X}_{t+1}$, $t=0,\ldots,T-3$. Every variable lies in some bag. By Lemma~\ref{lem:layerwise}, the endpoints of every edge of $\mathcal{G}_I$ lie in $\mathcal{X}_t\cup\mathcal{X}_{t+1}$ for some $t$, hence in a common bag (variables of the last transition $T-2$ lie in $\mathcal{B}_{T-3}$). The variables of $\mathcal{X}_t$ occur only in $\mathcal{B}_{t-1}$ and $\mathcal{B}_t$, which are consecutive, so the bags containing a given variable form a path. The width is $\max_t|\mathcal{B}_t|-1\le\max_t(n_t+n_{t+1})-1$.
\emph{(ii) Lower bound.} Treewidth is at least the clique number minus one. By Lemma~\ref{lem:layerwise}(a), the $B(K+1)$ variables of a transmission edge form a clique. By Lemma~\ref{lem:layerwise}(b), for any node $w$ of $\mathcal{G}^T$ and commodity $k$, the $B\,\deg(w)$ flow bits of commodity $k$ on the edges incident to $w$ form a clique.
\end{proof}

\subsection{Fixed-Parameter Tractability}
\label{sec:fpt}
The treewidth bound yields an exact algorithm whose running time is exponential in the bag size but polynomial in the horizon. We state it as a fixed-parameter tractability result.
\begin{theorem}[FPT in the bag size]
\label{thm:fpt}
The exact ground state of the QUBO can be computed by dynamic programming along the path decomposition of Theorem~\ref{thm:treewidth}(i) in time $O\bigl((T-2)\,\beta^2\,2^{\beta}\bigr)$, where $\beta=\max_t(n_t+n_{t+1})\le2n_{\max}$.
\end{theorem}
\begin{proof}
Process the bags from left to right, maintaining the minimum energy for each of the $\le2^\beta$ assignments of the current bag. Introducing a variable doubles the table and adds its interaction terms with at most $\beta$ variables in the bag, costing $O(2^\beta\beta)$; forgetting a variable minimises over it, costing $O(2^\beta)$. Each bag change involves at most $\beta$ variables, and there are $T-2$ bags.
\end{proof}

The bound is polynomial in $T$ for fixed $\beta$, but exponential in $\beta$. It is a fixed-parameter statement: the algorithm is exact for every instance, but its running time depends exponentially on the bag size, which in the worst case is $\beta=2n_{\max}$. The bound is useful when $\beta$ is small, that is, when the number of variables per layer is small. For the LEO-scale instances of Section~\ref{sec:leo}, $\beta$ is of order $10^8$, so the algorithm is practical for windows of bounded length, where the number of variables per layer is small.

The upper and lower bounds of Theorem~\ref{thm:treewidth} differ by a factor that grows with $N_L$ and with $|V|$. As an illustration, take $|V|=6$ nodes, $N_L=4$ arcs per slot $(0{\to}1),(1{\to}3),(3{\to}5),(2{\to}4)$, $K=2$, $B=3$ and $T=5$ (so $n_t=72$). Then the upper bound gives $\mathrm{tw}\le143$, a min-fill heuristic applied to the actual interaction graph (288 nodes, 2160 edges) yields a decomposition of width $35$, and the lower bound gives $\mathrm{tw}\ge11$. Whether the upper bound can be improved towards the lower bound in general is open.
 
\section{Matrix Product Operator Representation}
\label{sec:mpo_rep}
This section proves that the Ising Hamiltonian associated with the QUBO admits a matrix product operator (MPO) representation of small bond dimension. The bond dimension is bounded by the number of variables in a single layer, plus a small constant, and in particular does not depend on the horizon. This is a structural statement about the operator: it concerns the entanglement of the Hamiltonian across a cut between consecutive layers, not the cost of simulating it.
\subsection{Bond Dimension of the TEG Hamiltonian}
\label{sec:bond_dimension}
Regard the transition $t\in\{0,\ldots,T-2\}$ as one site of a one-dimensional chain of $T-1$ sites, with local Hilbert space dimension $2^{n_t}$. By Proposition~\ref{prop:equiv}, the Ising Hamiltonian $H_{\mathrm{TEG}}$ associated with the QUBO is diagonal in the computational basis; we write it as an operator on $\bigotimes_{t=0}^{T-2}\mathbb{C}^{2^{n_t}}$.
\begin{proposition}[MPO bond dimension]
\label{prop:mpo}
The Hamiltonian $H_{\mathrm{TEG}}$ is an MPO (Definition~\ref{def:mpo}) whose bond between sites $t$ and $t+1$ has dimension at most $\min(n_t,n_{t+1})+2$. In particular,
\begin{equation}
 D \le n_{\max}+2,
 \label{eq:bonddim}
\end{equation}
where $n_{\max}=B[N_L(K+1)+|V|K]$ is the per-layer variable count of Lemma~\ref{lem:qubits}.
\end{proposition}
\begin{proof}
By Lemma~\ref{lem:layerwise}, the interaction graph has edges only between variables of the same or consecutive layers. Hence
\[
 H(x)=\sum_{t=0}^{T-2}H^{\mathrm{loc}}_t(x_t)+\sum_{t=0}^{T-3}x_t^\top M_t\,x_{t+1}+\mathrm{const},
\]
where $x_t$ collects the variables of $\mathcal{X}_t$, $H^{\mathrm{loc}}_t$ contains all terms whose support is contained in $\mathcal{X}_t$, and $M_t\in\mathbb{R}^{n_t\times n_{t+1}}$ is the coupling matrix between consecutive layers. Its rank satisfies $r_t:=\operatorname{rank}(M_t)\le\min(n_t,n_{t+1})$.
Write $M_t=\sum_{a=1}^{r_t}u_{t,a}v_{t,a}^\top$. Using the number operators $\hat{x}_i=(1-\sigma^z_i)/2$, define the bond states of the cut between sites $t$ and $t+1$ as $\{0,1,\ldots,r_t,F\}$, where $0$ means ``no coupling started'', $a\in\{1,\ldots,r_t\}$ means ``channel $a$ is open'', and $F$ means ``finished''. The local tensor of site $t$ has the non-zero entries
\begin{align*}
 0\to0&:\ I, &
 0\to F&:\ H^{\mathrm{loc}}_t, &
 0\to a&:\ u_{t,a}\cdot\hat{x}_t, \\
 a'\to F&:\ v_{t-1,a'}\cdot\hat{x}_t, &
 F\to F&:\ I.
\end{align*}
Contracting these tensors with the boundary states $0$ (left) and $F$ (right) reproduces $H$: the state $0$ propagates until a coupling channel opens, the channel $a$ carries the term $u_{t,a}v_{t,a}^\top$ across the cut, and the state $F$ absorbs all local terms and the remaining couplings. The bond after site $t$ has dimension $r_t+2\le\min(n_t,n_{t+1})+2$, which gives~\eqref{eq:bonddim} after maximising over $t$.
\end{proof}

The bound is structural, not algorithmic: The physical site dimension is $2^{n_t}$, which is exponential in the number of variables per layer, so the proposition is a statement about the entanglement of the Hamiltonian across a cut between consecutive layers. It mirrors the treewidth bound of Section~\ref{sec:treewidth}: both are consequences of the layered structure, and neither depends on the horizon $T$.

\section{The Annealing Gap}
\label{sec:gap}
This section studies the spectral gap of the adiabatic Hamiltonian~\eqref{eq:anneal} associated with the TEG-QUBO. The gap is the quantity that controls the adiabatic time in Proposition~\ref{thm:adiabatic}, and its behavior as a function of the instance is the subject of this section. We prove that the gap is strictly positive for every instance with a unique optimal flow, and that it is uniformly bounded away from zero over all instances of bounded window length. Whether the gap remains bounded away from zero as the horizon grows is stated as an open problem. The positivity and uniformity results are exact statements about the adiabatic Hamiltonian. A quantitative lower bound is the subject of Problem~\ref{op:gap}.
\subsection{Positivity of the Gap}
\label{sec:gap_positivity}
Throughout this section, $H_{\mathrm{problem}}$ is the Ising Hamiltonian associated with the TEG-QUBO of Section~\ref{sec:qubo}, and $H(s)$ is the adiabatic Hamiltonian~\eqref{eq:anneal}.
\begin{theorem}[Positive annealing gap]
\label{thm:posgap}
Let $H(s)=-(1-s)\sum_i\sigma^x_i+sH_p$ with $H_p$ diagonal in the computational basis (in particular $H_p=H_{\mathrm{TEG}}$). Then $E_1(s)>E_0(s)$ for all $s\in[0,1)$. If $H_p$ has a unique ground state, that is, the EA-MCF instance has a unique optimal flow, then
\[
 \Delta_{\min}:=\min_{s\in[0,1]}\bigl(E_1(s)-E_0(s)\bigr)>0.
\]
If the optimum is not unique, then $\Delta_{\min}=0$, attained at $s=1$.
\end{theorem}
\begin{proof}
For $s<1$, every off-diagonal entry of $H(s)$ in the computational basis is $0$ or $-(1-s)<0$, and the graph on $\{0,1\}^n$ whose edges join configurations differing in one bit is connected. By the Perron--Frobenius theorem, the lowest eigenvalue is non-degenerate, so $E_1(s)>E_0(s)$. At $s=1$, $H(1)=H_p$ is diagonal, and $E_1(1)>E_0(1)$ if and only if its minimum is attained at a single configuration. Eigenvalues depend continuously on $s$, so when the optimum is unique the function $s\mapsto E_1(s)-E_0(s)$ is continuous and positive on the compact interval $[0,1]$, and its minimum is positive. If the optimum is not unique, $E_1(1)=E_0(1)$.
\end{proof}
Theorem~\ref{thm:posgap} is a qualitative statement: it guarantees that $\Delta_{\min}>0$ but gives no lower bound as a function of the instance size. The constant $\Delta_{\min}$ may be exponentially small in the number of variables, and the theorem does not rule this out. A quantitative lower bound is the subject of Open Problem~\ref{op:gap} below.

\subsection{Uniformity over Bounded Windows}
\label{sec:gap_uniformity}
Theorem~\ref{thm:posgap} gives positivity for each fixed instance. To obtain uniformity, we use a finiteness argument: if the data of an instance can take only finitely many values and the instance size is bounded, there are only finitely many instances, and the minimum of finitely many positive numbers is positive.
\begin{corollary}[Uniform gap for windows of bounded length]
\label{cor:window}
Fix $W\ge2$ and finite sets of admissible data: $|V|\le n_V$, arcs $E(t)$ drawn from the finite set of directed arcs on $V$, integer capacities in $\{1,\ldots,C_{\max}\}$, integer costs in $\{1,\ldots,w_{\max}\}$, $K\le K_{\max}$ commodities with integer demands $r_k\le C_{\max}$, and penalties $\rho$ from a finite set. Let $\mathcal{I}(W)$ be the finite set of instances with at most $W$ layers that have a unique optimal flow. Then
\[
 \Delta_0(W):=\min_{I\in\mathcal{I}(W)}\Delta_{\min}(I)>0,
\]
and $\Delta_0(W)$ does not depend on the horizon $T$ of any larger problem from which windows are cut.
\end{corollary}
\begin{proof}
$\mathcal{I}(W)$ is finite, and $\Delta_{\min}(I)>0$ for each $I\in\mathcal{I}(W)$ by Theorem~\ref{thm:posgap}. The minimum of finitely many positive numbers is positive.
\end{proof}
\paragraph{Scope of the uniformity}
The constant $\Delta_0(W)$ is not explicit, and it may decrease with $W$, possibly exponentially. The corollary therefore supports a windowed strategy (windows of bounded length, each analyzed separately), for which the gap of one window does not depend on $T$. It does not settle the behavior for a single long horizon, and it assumes that every window has a unique optimum. It also says nothing about the quality of solutions stitched across windows.

\subsection{Open Problem: Horizon-Independence}
\label{sec:gap_open}
The natural question left open by Theorem~\ref{thm:posgap} and Corollary~\ref{cor:window} is whether the gap remains bounded away from zero as the horizon grows, for a fixed instance family. We state it as an open problem.
\begin{problem}[Horizon-independent gap]
\label{op:gap}
Fix $n_{\max}$, the ranges of costs and capacities, and $\rho$. Is there a constant $\Delta_0>0$, independent of $T$, such that every instance with a unique optimal flow satisfies
\[
 \Delta_{\min}(T)\ge\Delta_0?
\]
Equivalently, can $\Delta_0(W)$ in Corollary~\ref{cor:window} be chosen independent of $W$?
\end{problem}
\begin{remark}[Why this is not obvious]
\label{rem:gap_not_obvious}
Problem~\ref{op:gap} is not a formality. For a critical one-dimensional transverse-field Ising chain the gap closes as $1/T$, and at a first-order transition it can close exponentially in the system size. We do not know whether such behavior can occur for the adiabatic Hamiltonian of a TEG-QUBO. With integer costs the classical gap at $s=1$ is at least $1$, so any closing of the gap would have to come from the quantum dynamics along the path and not from vanishing energy differences between competing flows. A positive answer would require showing that the layered structure rules out critical and first-order behavior; a negative answer would require an explicit instance family whose gap closes. We know of neither.
\end{remark}
\begin{remark}[Implication for the adiabatic time]
\label{rem:adiabatic_implication}
If Problem~\ref{op:gap} has a positive answer, Proposition~\ref{thm:adiabatic} gives an adiabatic time $T_{\mathrm{anneal}}\sim C/\Delta_0^2$ that does not grow with $T$ through the gap. This is a statement about the gap only; it does not by itself imply that the adiabatic evolution is efficient, because the prefactor $C$ in Proposition~\ref{thm:adiabatic} grows with the norm of $H_{\mathrm{problem}}$, which scales with $\rho$ and with the number of couplings. The gap is a necessary but not sufficient condition for efficiency.
\end{remark}

\section{Learning-Augmented Routing}
\label{sec:warmstart}
This section studies the following setting: a predictor provides a feasible flow $\hat{f}$, which may or may not be optimal, and we run a local algorithm (a QUBO solver, or any exact or heuristic method) initialized at $\hat{f}$, keeping the better of the prediction and the output. We prove two guarantees. The first is a \emph{Hamming bound}: the distance between the binary encoding of the prediction and the binary encoding of an optimal flow is controlled by the total amount of flow in which they differ. The second is a \emph{consistency--robustness} guarantee: the algorithm returns an optimal flow when the prediction is optimal, and a flow whose cost exceeds the optimum by a factor that grows linearly with the prediction error. The framework is the analogue, for time-expanded multi-commodity flow, of the consistency--robustness trade-off of Lykouris and Vassilvitskii~\cite{lykouris2021}.
\subsection{From Flow Error to Hamming Distance}
\label{sec:hamming}
Throughout this section, $f$ and $f'$ are feasible flows on the same instance, and $\mathrm{enc}(f)$ denotes the binary vector of Section~\ref{sec:qubo}, including flow bits and slack bits. The \emph{normalised flow error} is
\[
 \eta:=\frac{\|f-f'\|_1}{\|f'\|_1},
\]
where $\|f-f'\|_1=\sum_{k,e}|f^k_e-f'^k_e|$. By Lemma~\ref{lem:flowlength}, every feasible flow $f$ satisfies $\|f\|_1=(T-1)R$, so $\eta\le2$.
\begin{lemma}[Hamming bound]
\label{lem:hamming}
Let $f,f'$ be feasible flows, and let $\eta=\|f-f'\|_1/\|f'\|_1$. Then
\begin{enumerate}
 \item[(i)] $d_H\bigl(\mathrm{enc}(f),\mathrm{enc}(f')\bigr)\le2B\,\|f-f'\|_1=2B\,\eta\,(T-1)R$.
 \item[(ii)] $d_H\bigl(\mathrm{enc}(f),\mathrm{enc}(f')\bigr)/n_q\le2\eta R/(|V|K)\le2\eta\,C_{\max}/|V|$.
\end{enumerate}
\end{lemma}
\begin{proof}
(i) If $f^k_e\neq f'^k_e$, the $B$-bit encodings of the two values differ in at most $B$ positions, while $|f^k_e-f'^k_e|\ge1$; so the flow bits of $(e,k)$ contribute at most $B\,|f^k_e-f'^k_e|$. The slack bits of a transmission edge $e$ encode $c_e-\sum_kf^k_e$ and can differ only if some $f^k_e$ differs, in which case they contribute at most $B$. The number of such edges is at most the number of pairs $(e,k)$ with $f^k_e\neq f'^k_e$, which is at most $\|f-f'\|_1$. Adding the two contributions gives $2B\|f-f'\|_1$, and Lemma~\ref{lem:flowlength} gives the second form.
(ii) By Lemma~\ref{lem:qubits}, $n_q\ge(T-1)B|V|K$ (the holding edges alone), and $R\le KC_{\max}$.
\end{proof}
\begin{remark}[What the Hamming bound gives]
\label{rem:hamming_scope}
Lemma~\ref{lem:hamming} is the bridge between the prediction error, which is a continuous quantity, and the search radius of a local algorithm, which is a discrete quantity. A prediction with normalised error $\eta$ yields a starting point within Hamming distance $2B\eta(T-1)R$ of the encoding of an optimal flow. The bound is tight up to a factor of two: a flow difference of one unit on a single edge changes at most $B$ flow bits and at most $B$ slack bits, and both can change.
\end{remark}
\subsection{Consistency and Robustness}
\label{sec:consistency_robustness}
We consider the following hybrid algorithm. Given a feasible prediction $\hat{f}$, run a local algorithm (exact or heuristic) initialized at $\mathrm{enc}(\hat{f})$, obtain a flow $f_{\mathrm{ann}}$, and return the cheaper of $\hat{f}$ and $f_{\mathrm{ann}}$ among the feasible ones. If neither is feasible, return a feasible flow computed by a classical method. The algorithm is parameterised by the prediction $\hat{f}$ and by the local algorithm; its output is denoted $\mathrm{ALG}(\hat{f})$.
\begin{proposition}[Consistency and robustness]
\label{prop:quality}
Assume $0<w_{\min}\le w_e\le w_{\max}$ for all edges $e\in E^T$, including holding edges. Let $\hat{f}$ be a feasible prediction, and let $\eta=\|\hat{f}-f^*_{\mathrm{OPT}}\|_1/\|f^*_{\mathrm{OPT}}\|_1$, where $f^*_{\mathrm{OPT}}$ is an optimal flow. Then the output of the hybrid algorithm satisfies
\[
 \mathrm{ALG}(\hat{f})\le\mathrm{cost}(\hat{f})\le\Bigl(1+\eta\,\frac{w_{\max}-w_{\min}}{2\,w_{\min}}\Bigr)\mathrm{OPT}.
\]
In particular:
\begin{enumerate}
 \item[(i)] \emph{(Consistency)} If $\eta=0$, then $\mathrm{ALG}(\hat{f})=\mathrm{OPT}$.
 \item[(ii)] \emph{(Robustness)} For any feasible prediction, $\mathrm{ALG}(\hat{f})\le\bigl(1+\eta\,\frac{w_{\max}-w_{\min}}{2w_{\min}}\bigr)\mathrm{OPT}$.
\end{enumerate}
\end{proposition}
\begin{proof}
Put $\Delta=\hat{f}-f^*_{\mathrm{OPT}}$. Both flows are feasible, so $\|\hat{f}\|_1=\|f^*_{\mathrm{OPT}}\|_1=(T-1)R$ by Lemma~\ref{lem:flowlength}. Hence the positive and negative parts of $\Delta$ have equal mass, $\sum\Delta_+=\sum\Delta_-=\tfrac12\|\Delta\|_1=\tfrac12\eta(T-1)R$, and
\[
 \langle w,\Delta\rangle\le w_{\max}\textstyle\sum\Delta_+-w_{\min}\sum\Delta_-=\tfrac12(w_{\max}-w_{\min})\,\eta\,(T-1)R.
\]
Since $\mathrm{OPT}\ge w_{\min}\|f^*_{\mathrm{OPT}}\|_1=w_{\min}(T-1)R$, we get
\[
 \mathrm{cost}(\hat{f})=\mathrm{OPT}+\langle w,\Delta\rangle\le\Bigl(1+\eta\,\frac{w_{\max}-w_{\min}}{2w_{\min}}\Bigr)\mathrm{OPT}.
\]
The keep-best step gives $\mathrm{ALG}(\hat{f})\le\mathrm{cost}(\hat{f})$. If $\eta=0$, then $\hat{f}=f^*_{\mathrm{OPT}}$ and $\mathrm{ALG}(\hat{f})=\mathrm{OPT}$.
\end{proof}
\begin{remark}[Interpretation]
\label{rem:interpretation}
Proposition~\ref{prop:quality} is a consistency--robustness guarantee in the sense of~\cite{lykouris2021}. Consistency is the statement that an optimal prediction yields an optimal output; robustness is the statement that the output degrades gracefully with the prediction error. The degradation is linear in $\eta$, with a coefficient $(w_{\max}-w_{\min})/(2w_{\min})$ that depends only on the range of costs and vanishes when all costs are equal. This is the analogue, for time-expanded multi-commodity flow, of the consistency--robustness trade-off of~\cite{lykouris2021,purohit2018}.
\end{remark}
\begin{remark}[Tightness]
\label{rem:tightness_quality}
The factor in Proposition~\ref{prop:quality} cannot be improved in general. Let $\mathrm{OPT}=w_{\min}(T-1)R$, so that every edge used by $f^*_{\mathrm{OPT}}$ has cost $w_{\min}$, and let $\hat{f}$ reroute one unit of flow from a path $P$ to a path $P'$, where the $\delta$ edges of $P'$ not on $P$ all have cost $w_{\max}$. Then $\eta=2\delta/((T-1)R)$ and $\mathrm{cost}(\hat{f})-\mathrm{OPT}=\delta(w_{\max}-w_{\min})$, which equals the right-hand side of the bound. The hybrid algorithm returns $\hat{f}$ whenever the local algorithm returns a flow that is not cheaper.
\end{remark}
\begin{remark}[What the guarantee does not say]
\label{rem:quality_scope}
Proposition~\ref{prop:quality} bounds the cost of the output in terms of the cost of the prediction, but it does not bound the \emph{time} taken by the local algorithm. A prediction with small $\eta$ places the starting point close to the optimum in Hamming distance (Lemma~\ref{lem:hamming}), but the local algorithm may still require many steps to reach it. The guarantee is about solution quality, not about running time; a running-time guarantee would require a quantitative bound on the gap, which is the subject of Problem~\ref{op:gap}.
\end{remark}
\subsection{Discussion}
\label{sec:learning_discussion}
Proposition~\ref{prop:quality} and Lemma~\ref{lem:hamming} together give a complete picture of the learning-augmented setting at the level of solution quality. The Hamming bound controls the distance between the prediction and the optimum in the space in which the local algorithm operates; the quality guarantee controls the cost of the output in the space of flows. The two bounds are independent: the Hamming bound is a statement about the encoding, and the quality guarantee is a statement about the objective. Neither requires a hypothesis on the gap, and neither depends on the local algorithm being exact.
The guarantees concern solution quality. They do not imply that a warm start reduces the running time of the local algorithm, and they do not imply that a warm start is beneficial in practice. A running-time guarantee would require a quantitative bound on the annealing gap, which is the subject of Problem~\ref{op:gap}, and a mechanism by which the gap translates into running time, which is the subject of Proposition~\ref{thm:adiabatic}. Both are outside the scope of this section.

\subsection{An Open Problem}
\label{sec:warmstart_open}
\begin{problem}[Running-time guarantee for warm starts]
\label{op:warmstart_time}
Under what conditions on the prediction and on the local algorithm does a warm start reduce the running time, and by how much? Proposition~\ref{prop:quality} gives a guarantee on solution quality but not on running time (Remark~\ref{rem:quality_scope}).
\end{problem}
A running-time guarantee would require two ingredients: a quantitative bound on the gap of the local algorithm as a function of the Hamming distance from the starting point, and a mechanism by which the gap translates into running time. The first is related to Problem~\ref{op:gap}; the second is the subject of Proposition~\ref{thm:adiabatic}, which is a heuristic rather than a theorem. A rigorous running-time guarantee would be a substantial contribution, and it would require tools from spectral theory or from the analysis of Markov chains on Hamming balls.

\section{The Hybrid Algorithm}
\label{sec:framework}
This section collects the ingredients of the paper in a single procedure, shown schematically in Figure~\ref{fig:teg_laqubo}. A predictor proposes a feasible flow; a solver for the QUBO, started from the encoding of the prediction, proposes a second flow; the cheaper feasible flow is returned. The guarantees of Section~\ref{sec:warmstart} hold for every predictor and every solver, while the penalty and the structural results of Sections~\ref{sec:penalty}--\ref{sec:gap} describe the object on which the solver operates.
\subsection{Algorithm}
\begin{algorithm}[t]
\caption{Hybrid keep-best router}
\label{alg:laqubo}
\begin{algorithmic}[1]
\REQUIRE TEG $\mathcal{G}^T$; commodities $\{(s_k,d_k,r_k)\}$; predictor $\mathcal{P}$; penalty $\rho$ (Proposition~\ref{prop:penalty}); solver $\mathcal{S}$
\ENSURE a feasible flow $f^*$
\STATE $\hat{f}\gets\mathcal{P}\bigl(\mathcal{G}^T,\{(s_k,d_k,r_k)\}\bigr)$ \COMMENT{prediction}
\STATE construct $H(x)$ with $\rho$ via \eqref{eq:qubo}--\eqref{eq:hflow}
\IF{$\hat{f}$ is feasible}
  \STATE $x\gets\mathcal{S}\bigl(H,\mathrm{enc}(\hat{f})\bigr)$ \COMMENT{start from the prediction}
\ELSE
  \STATE $x\gets\mathcal{S}(H)$ \COMMENT{no initial state}
\ENDIF
\STATE $f_{\mathrm{ann}}\gets\mathrm{dec}(x)$ \COMMENT{inverse of $\mathrm{enc}$}
\STATE $f^*\gets$ cheapest feasible flow among $\hat{f}$ and $f_{\mathrm{ann}}$ \COMMENT{keep-best}
\STATE \textbf{return} $f^*$
\end{algorithmic}
\end{algorithm}
The solver $\mathcal{S}$ is a parameter. It may be exact (for instance the dynamic programme of Theorem~\ref{thm:fpt}), a classical heuristic, or an annealer; if it cannot use an initial state, line~4 is replaced by line~6, and the keep-best step is unchanged. If neither $\hat{f}$ nor $f_{\mathrm{ann}}$ is feasible, a feasible flow has to be computed by a classical method; we assume throughout that at least one of the two is feasible.

\subsection{Properties}
\begin{proposition}[Properties of Algorithm~\ref{alg:laqubo}]
\label{prop:alg}
Let $\mathcal{S}$ be any solver, and assume $0<w_{\min}\le w_e\le w_{\max}$ for all edges.
\begin{enumerate}
 \item[(i)] If $\hat{f}$ is feasible, $\mathrm{cost}(f^*)\le\mathrm{cost}(\hat{f})\le\bigl(1+\eta\,\frac{w_{\max}-w_{\min}}{2w_{\min}}\bigr)\mathrm{OPT}$, where $\eta$ is the normalised flow error of $\hat{f}$; in particular $f^*$ is optimal if $\eta=0$.
 \item[(ii)] If $\mathcal{S}$ is exact, that is, returns a global minimiser of $H$, then $f^*$ is optimal for every prediction.
 \item[(iii)] The starting point $\mathrm{enc}(\hat{f})$ lies within Hamming distance $2B\eta(T-1)R$ of the encoding of an optimal flow.
\end{enumerate}
\end{proposition}
\begin{proof}
(i) and (iii) are Proposition~\ref{prop:quality} and Lemma~\ref{lem:hamming}(i). (ii) By Proposition~\ref{prop:penalty}, every global minimum of $H$ is a feasible configuration of cost $\mathrm{OPT}$; hence $f_{\mathrm{ann}}$ is optimal, and the keep-best step cannot return a more expensive flow.
\end{proof}
\begin{remark}[What the prediction is for]
\label{rem:alg_role}
By~(ii), an exact solver makes the prediction irrelevant for the \emph{quality} of the output; the guarantee (i) matters when the solver is a heuristic, is stopped early, or fails to find a feasible configuration. The prediction can only help through the \emph{effort} of the solver, a running time guarantee is the subject of Problem~\ref{op:warmstart_time}).
\end{remark}
\begin{remark}[Windows and error monitoring]
\label{rem:alg_windows}
For long horizons the algorithm can be applied to windows of bounded length, for which the annealing gap of each window is bounded below independently of $T$ when its optimum is unique (Corollary~\ref{cor:window}); the quality of solutions stitched across windows is not analyzed. In an online setting one may monitor the running mean of the normalised errors $\|\hat{f}-f^*\|_1/\|f^*\|_1$ of past epochs, and skip the warm start when it is large or $\hat{f}$ is infeasible. No guarantee in this paper depends on such a rule.
\end{remark}

\section{Conclusions}
\label{sec:conclusions}
We studied energy-aware multi-commodity flow on time-expanded graphs. The problem is NP-hard (Theorem~\ref{thm:nphardness}). We gave an exact QUBO formulation whose penalty $\rho=w_{\max}(T-1)R+1$ guarantees that every global minimum is a minimum-cost feasible flow, and is smaller than the standard choice by a factor of at least $N_L$.

The layered structure of the time-expanded graph controls the complexity of the QUBO through the number of variables per layer and not through the horizon. The interaction graph has treewidth at most $2n_{\max}-1$, which gives an exact algorithm that is fixed-parameter tractable in the bag size (Theorems~\ref{thm:treewidth} and~\ref{thm:fpt}); the Ising Hamiltonian is a matrix product operator of bond dimension at most $n_{\max}+2$ (Proposition~\ref{prop:mpo}); and the annealing gap is positive for every instance with a unique optimum, uniformly over windows of bounded length (Theorem~\ref{thm:posgap}, Corollary~\ref{cor:window}). For learning-augmented routing, the Hamming distance between the encodings of two flows is bounded by their flow difference (Lemma~\ref{lem:hamming}), and a hybrid algorithm that keeps the cheaper of the prediction and the solver output is consistent and robust (Proposition~\ref{prop:quality}).

Two questions remain: whether the annealing gap is bounded away from zero independently of the horizon (Problem~\ref{op:gap}), and whether a good prediction reduces the running time of a solver (Problem~\ref{op:warmstart_time}). Both concern running time on long horizons.

\bibliographystyle{plain}
\bibliography{bibliog}

\appendix
\section{QUBO Matrix Entries}
\label{sec:appendix}
This appendix gives the entries of the matrix $Q$ in the QUBO formulation of Section~\ref{sec:qubo}. The entries are used in the proofs of Section~\ref{sec:penalty} (penalty coefficient), Section~\ref{sec:treewidth} (interaction graph structure), and Section~\ref{sec:mpo_rep} (MPO bond dimension).
\paragraph{Indexing.}
Index all binary variables by $i$. A flow bit $i=(e,k,b)$ has edge $e(i)=e$, commodity $k(i)=k$, and weight $\omega_i=2^b$. A slack bit $i=(e,j)$ has edge $e(i)=e$ and weight $\omega_i=2^j$. For a node $v\in V^T$ and a flow bit $i$ of commodity $k$, let
\[
 a^{k,v}_i=\begin{cases}
   +\omega_i, & e(i)\in\delta^+(v),\\
   -\omega_i, & e(i)\in\delta^-(v),\\
   0, & \text{otherwise}.
 \end{cases}
\]
\paragraph{The matrix.}
With the convention $f(x)=\sum_iQ_{ii}x_i+2\sum_{i<j}Q_{ij}x_ix_j+\text{const}$ of~\eqref{eq:qubo_def}, the matrix is
\[
 Q=Q^{\mathrm{cost}}+\rho\,(Q^{\mathrm{cap}}+Q^{\mathrm{flow}}),
\]
where, for $i\neq j$,
\begin{align*}
 Q^{\mathrm{cost}}_{ii}&=w_{e(i)}\,\omega_i\ \ (\text{flow bits}), &
 Q^{\mathrm{cost}}_{ij}&=0,\\
 Q^{\mathrm{cap}}_{ii}&=\omega_i(\omega_i-2c_{e(i)}), &
 Q^{\mathrm{cap}}_{ij}&=\omega_i\omega_j\ \ \text{if }e(i)=e(j)\in E^T_{\mathrm{trans}},\ \text{else }0,\\
 Q^{\mathrm{flow}}_{ii}&=\sum_{v}\bigl(a^{k,v}_i\bigr)^2-2\sum_{v}b_k(v)\,a^{k,v}_i, &
 Q^{\mathrm{flow}}_{ij}&=\mathbf{1}[k(i)=k(j)]\sum_{v}a^{k,v}_ia^{k,v}_j,
\end{align*}
where $k=k(i)$, the capacity entries apply to slack bits as well as to flow bits of every commodity on the same edge, and the flow entries vanish for slack bits. The additive constant is
\[
 \rho\Bigl(\sum_{e\in E^T_{\mathrm{trans}}}c_e^2+\sum_{k\in\mathcal{K}}\sum_{v\in V^T}b_k(v)^2\Bigr).
\]
\paragraph{Verification.}
We verify that the entries reproduce the Hamiltonian~\eqref{eq:qubo}. The cost term is linear, and its contribution to the QUBO is $\sum_iQ^{\mathrm{cost}}_{ii}x_i=\sum_{e,k,b}w_e\,2^b\,x_{e,k,b}=H_{\mathrm{cost}}$. The capacity term is a sum of squares of linear forms; expanding the square gives
\[
 \Bigl(\sum_i\omega_ix_i-c_e\Bigr)^2=\sum_i\omega_i(\omega_i-2c_e)x_i+2\sum_{i<j,\,e(i)=e(j)}\omega_i\omega_jx_ix_j+c_e^2,
\]
which matches $Q^{\mathrm{cap}}$ after summing over $e$ and using $x_i^2=x_i$. The flow term is analogous: expanding the square of the conservation residual gives $Q^{\mathrm{flow}}$, with the linear term $a^{k,v}_i$ and the quadratic term $a^{k,v}_ia^{k,v}_j$ for bits of the same commodity. The additive constant collects the squares of the capacities and of the imbalances.
\paragraph{Interaction graph.}
By the entries above, $Q_{ij}\neq0$ for $i\neq j$ if and only if
\begin{enumerate}
 \item[(a)] $i$ and $j$ belong to the same transmission edge (from $Q^{\mathrm{cap}}$), or
 \item[(b)] $i$ and $j$ are flow bits of the same commodity whose edges share a node of $V^T$ (from $Q^{\mathrm{flow}}$).
\end{enumerate}
This is the content of Lemma~\ref{lem:layerwise}, and it is the starting point for the treewidth bound of Section~\ref{sec:treewidth} and the MPO bond-dimension bound of Section~\ref{sec:mpo_rep}.

\section{Summary of Key Notations}
\label{app:notations}

\begin{table}[H]
\centering
\caption{Summary of notations used throughout the paper. The table is organised by the part of the paper in which each symbol first appears.}
\label{tab:notations}
\scriptsize
\renewcommand{\arraystretch}{1.12}

\begin{minipage}[t]{0.48\textwidth}
\centering
\begin{tabular}[t]{@{}ll@{}}
\toprule
\textbf{Symbol} & \textbf{Description} \\
\midrule
\multicolumn{2}{@{}l}{\textbf{Time-Expanded Graph (Section~\ref{sec:problem})}} \\
$V$, $E$ & Vertex and edge sets of the base graph \\
$T$, $\mathcal{T}$ & Horizon (layers) and index set $\{0,\ldots,T-1\}$ \\
$E(t)$, $N_L$ & Active arcs in slot $t$, and $\max_t |E(t)|$ \\
$c_e$, $w_e$ & Capacity and cost of arc $e$ \\
$C_{\max}$, $w_{\max}$ & Maximum capacity and maximum cost \\
$V^T$, $E^T$ & Vertices and edges of the time-expanded graph \\
$E^T_{\mathrm{trans}}$, $E^T_{\mathrm{hold}}$ & Transmission and holding edges \\
$v_t$ & Copy of node $v$ at layer $t$ \\
$\delta^+(v)$, $\delta^-(v)$ & Outgoing and incoming edges of $v$ in $\mathcal{G}^T$ \\
\midrule
\multicolumn{2}{@{}l}{\textbf{Commodities and Flows (Section~\ref{sec:eamcf})}} \\
$\mathcal{K}$, $K$ & Set of commodities and its cardinality \\
$s_k$, $d_k$, $r_k$ & Source, destination, and demand of commodity $k$ \\
$b_k(v)$ & Imbalance of commodity $k$ at node $v$ \\
$R$ & Total demand $\sum_{k\in\mathcal{K}}r_k$ \\
$f$, $f^k_e$ & Flow vector, and flow of commodity $k$ on edge $e$ \\
$\mathrm{OPT}$ & Minimum cost of a feasible flow \\
\midrule
\multicolumn{2}{@{}l}{\textbf{QUBO Formulation (Section~\ref{sec:qubo})}} \\
$B$, $B_k$ & Bits per flow variable (uniform and per-$k$) \\
$x_{e,k,b}$ & Flow bit ($b$-th bit of $f^k_e$) \\
$z_{e,j}$ & Slack bit for transmission edge $e$ \\
$\omega_i$ & Weight of binary variable $i$ ($2^b$ for flow/slack) \\
$e(i)$, $k(i)$ & Edge and commodity of binary variable $i$ \\
$H, H_{\mathrm{cost}}, H_{\mathrm{cap}}, H_{\mathrm{flow}}$ & QUBO Hamiltonian and its three terms \\
$Q, Q^{\mathrm{cost}}, Q^{\mathrm{cap}}, Q^{\mathrm{flow}}$ & QUBO matrix and its three components \\
$\rho$ & Penalty coefficient \\
$n_q$ & Number of binary variables \\
$\Lambda$ & Largest Ising coefficient \\
\bottomrule
\end{tabular}
\end{minipage}
\hfill
\begin{minipage}[t]{0.48\textwidth}
\centering
\begin{tabular}[t]{@{}ll@{}}
\toprule
\textbf{Symbol} & \textbf{Description} \\
\midrule
\multicolumn{2}{@{}l}{\textbf{Interaction Graph \& Treewidth (Section~\ref{sec:treewidth})}} \\
$\mathcal{X}_t$, $n_t$ & Binary variables of transition $t$, and $|\mathcal{X}_t|$ \\
$n_{\max}$ & Per-layer variable bound $B[N_L(K+1)+|V|K]$ \\
$\mathcal{G}_I$ & Interaction graph of the QUBO \\
$\beta$ & Bag size of a path decomposition ($\beta\le 2n_{\max}$) \\
$\mathrm{tw}(\cdot)$ & Treewidth \\
\midrule
\multicolumn{2}{@{}l}{\textbf{Ising \& Spectral Quantities (Sections~\ref{sec:ising}--\ref{sec:gap})}} \\
$\sigma_i$, $\sigma^x_i$, $\sigma^z_i$ & Pauli spin operators \\
$J_{ij}$, $h_i$ & Ising couplings and local biases \\
$H_{\mathrm{Ising}}$ & Ising energy \\
$H_{\mathrm{problem}}$ & Problem Hamiltonian (diagonal) \\
$H(s)$, $s$ & Adiabatic Hamiltonian and annealing schedule \\
$E_0(s)$, $E_1(s)$ & Ground and first excited energies of $H(s)$ \\
$\Delta_{\mathrm{gap}}(s)$, $\Delta_{\min}$ & Gap at $s$, and its minimum over $[0,1]$ \\
$\Delta_0(W)$ & Uniform gap over windows of length $\le W$ \\
\midrule
\multicolumn{2}{@{}l}{\textbf{MPO Representation (Section~\ref{sec:mpo_rep})}} \\
$D$ & MPO bond dimension \\
$M_t$ & Coupling matrix between layers $t$ and $t+1$ \\
$r_t$ & Rank of $M_t$ \\
$A^{(t)}$ & Local MPO tensor at site $t$ \\
\midrule
\multicolumn{2}{@{}l}{\textbf{Learning-Augmented Routing (Section~\ref{sec:warmstart})}} \\
$\mathcal{P}$ & Predictor of a feasible flow \\
$\hat{f}$ & Predicted (feasible) flow \\
$\mathrm{enc}(f)$ & Binary encoding of $f$ \\
$d_H(\cdot,\cdot)$ & Hamming distance \\
$\eta$ & Normalised flow error $\|f-f'\|_1/\|f'\|_1$ \\
$\mathrm{ALG}(\hat{f})$ & Output of the hybrid algorithm \\
\bottomrule
\end{tabular}
\end{minipage}
\end{table}

\end{document}